\documentclass[lettersize,journal]{IEEEtran}
\usepackage{amsmath,amsfonts}
\usepackage{algorithmic}
\usepackage{algorithm}
\usepackage{array}
\usepackage[caption=false,font=normalsize,labelfont=sf,textfont=sf]{subfig}
\usepackage{textcomp}
\usepackage{stfloats}
\usepackage{url}
\usepackage{booktabs}
\usepackage{multirow}
\usepackage{verbatim}
\usepackage{graphicx}
\usepackage{cite}
\usepackage{pifont}
\usepackage{tabularx}
\usepackage{makecell}
\begin{document}

\title{Siamese Meets Diffusion Network: SMDNet for Enhanced Change Detection in High-Resolution RS Imagery}

\author{Jia Jia, Geunho Lee, Zhibo Wang, Lyu Zhi, and Yuchu He
\thanks{Jia Jia is with the Department of Culture Technology(Artificial Intelligence Direction), Jeonju University, Jeonju 55069, Republic of Korea.(e-mail: kimmit628@jj.ac.kr)

Geunho Lee is with the Department of Artificial Intelligence, Jeonju University, Jeonju 55069, Republic of Korea. (e-mail: ghlee@jj.ac.kr).

Zhibo Wang and Lyu Zhi are with the Department of Graduate School of Artificial Intelligence, Jeonju University, Jeonju-si 55069, Republic of Korea.(e-mail: wang970128@jj.ac.kr; lyuzhi@jj.ac.kr).

Yuchu He is with the Department of Information Science and Technology College, Zhengzhou Normal University, Zhengzhou 450044, China. (e-mail: heyuchu@foxmail.com)}}



\maketitle

\begin{abstract}
Recently, the application of deep learning to change detection (CD) has significantly progressed in remote sensing images. In recent years, CD tasks have mostly used architectures such as CNN and Transformer to identify these changes. However, these architectures have shortcomings in representing boundary details and are prone to false alarms and missed detections under complex lighting and weather conditions. For that, we propose a new network, Siamese Meets Diffusion Network (SMDNet). This network combines the Siam-U2Net Feature Differential Encoder (SU-FDE) and the denoising diffusion implicit model to improve the accuracy of image edge change detection and enhance the model's robustness under environmental changes. First, we propose an innovative SU-FDE module that utilizes shared weight features to capture differences between time series images and identify similarities between features to enhance edge detail detection. Furthermore, we add an attention mechanism to identify key coarse features to improve the model's sensitivity and accuracy. Finally, the diffusion model of progressive sampling is used to fuse key coarse features, and the noise reduction ability of the diffusion model and the advantages of capturing the probability distribution of image data are used to enhance the adaptability of the model in different environments. Our method's combination of feature extraction and diffusion models demonstrates effectiveness in change detection in remote sensing images. The performance evaluation of SMDNet on LEVIR-CD, DSIFN-CD, and CDD datasets yields validated F1 scores of 90.99\%, 88.40\%, and 88.47\%, respectively. This substantiates the advanced capabilities of our model in accurately identifying variations and intricate details.
\end{abstract}


\section{Introduction}
\IEEEPARstart{R}{emote} sensing (RS) \cite{9254004, 9645230, 9635703, 9939092, 9956008} technology has advanced quickly as a result of the "Earth Observation System" projects initiated by numerous nations and international organizations worldwide, including Landsat, Gaofen, SPOT, RADARSAT, Sentinel, and other satellite efforts. 
RS image analysis \cite{hu2023bag, qiao2022novel}, particularly for change detection (CD), allows us to compare alterations in objects or phenomena over different time intervals within a consistent location. 
This has become an indispensable tool in a variety of fields, including environmental monitoring and conservation, disaster assessment, and urban development planning.
Moreover, by generating and harnessing vast amounts of data from change detection in remote sensing images, we can achieve a deeper semantic understanding across multiple disciplines. This not only enhances our ability to accurately portray the dynamic changes occurring on Earth, but also contributes significantly to our knowledge of the planet's evolving landscape.

Although CD technology in RS images \cite{hu2023robust, weng2023cross} has made rapid progress, many problems still need to be solved in practical applications. Current CD technology is dealing with the lack of information caused by the mutual occlusion of ground objects, spectral confusion caused by different objects (such as trees, shrubs, buildings, etc.) showing similar spectral characteristics \cite{chen2021remote}, and the combination of climate conditions, lighting, seasonal changes, etc. Many challenges still exist in identifying distractors to extract clear edge details\cite{liu2004analysis}. As such, reliably extracting segmentation maps before and after alterations remains difficult.

Traditionally, CD has evolved from manual visual analysis to using algebraic techniques to calculate image pixel differences\cite{bruzzone2000automatic}\cite{hussain2013change} to employing data reduction strategies such as principal component analysis (PCA) \cite{shi2020change}. As image resolution increases, deep learning can automatically extract complex features from big data, showing advantages in feature extraction and representation learning. Currently, CNN \cite{shen2021exploring} and Transformer \cite{shen2023git} in deep learning have become the mainstream methods to improve the accuracy and efficiency of RS image CD. However, there is still room for improvement in feature processing and detailed description.

The success of U-Net \cite{ronneberger2015u} in medical image segmentation has inspired its application in Remote Sensing (RS) image Change Detection (CD). The Efficient Multi-Resolution Network (EMRN) \cite{shen2021efficient} proposes a module for uniform dimensionality of multi-resolution features, addressing dimensional inconsistencies in images of varying resolutions.
The UNet++ \cite{peng2019end} method adopts an end-to-end encoder-decoder structure to mitigate information loss stemming from the generation of difference maps from deep features. In the realm of RS image CD, SNUNet-CD \cite{fang2021snunet} innovatively combines a Siamese network and UNet++ to fuse deep and shallow features through compact information transmission and an attention mechanism.
STANet \cite{chen2020spatial}, leveraging twin networks and attention mechanisms, enhances spatiotemporal relationship modeling. Despite these advancements, Convolutional Neural Networks (CNNs) still grapple with challenges in handling multi-temporal image fusion under dynamic conditions and in complex environments.

The Transformer architecture was originally used in natural language processing (NLP) and is now widely used in the CD field of RS. For example, the Bitemporal Image Transformer (BIT) \cite{rs12101662} combines CNN and Transformer to improve changed region recognition. 
PBSL \cite{shen2023pbsl} introduces a multimodal alignment approach to highlight relevant features and suppress irrelevant information.
Changeformer \cite{bandara2022transformer} combines a layered Transformer encoder and a lightweight decoder to capture the information required for accurate CD. SwinSUNet \cite{zhang2022swinsunet} combines Siamese structure and Swin Transformer to enhance global information capture and CD effects. CTS-Unet\cite{heidarycts-unet2023} combines Siamese U-shaped network and convolutional Transformer, focusing on low-resolution image and temporal relationship processing. Although the Transformer is effective in global information capture, it faces challenges of computational efficiency and long-distance dependency capture when processing large-scale remote sensing datasets. We discovered the advantages of Siamese networks in CNN and Transformer to effectively analyze image pairs and adapt to the complex spatiotemporal relationships in remote sensing image CD through a shared weight mechanism.

In addition, some new research directions have also received attention. Diffusion models(DM) \cite{ho2020denoising} \cite{ramesh2022hierarchical} \cite{FID} \cite{kawar2022denoising}\cite{saharia2022image} are powerful generative models that generate complex images and patterns by gradually reversing the process of noise added to the data. Their main advantage is the ability to create extremely realistic images due to their ability to process complex patterns and their powerful learning capabilities to simulate and reconstruct the subtle structure of the data. Unlike traditional generative models, DM are more robust in generating clarity and diversity, making them powerful tools in medicine\cite{kazerouni2022diffusion}, artistic creation, media production, and design. For example, virtual dress-up\cite{zhang2023two}, advancing Pose\cite{shen2023advancing}, and video generation\cite{ho2022imagen}\cite{wu2023tune}.

However, few related studies apply DM to RS image processing. Only a few works have recently applied the DM to RS image processing, which mainly inputs a large amount of collected RS image data into diffusion. The model acts as an extractor of key semantic features or as a pre-trained encoder\cite{gedara2022remote}. Inspired by the success of DDPM in multiple fields, we propose the SMDNet model to improve edge detection accuracy in CD and robustness under different environmental conditions. This model combines SU-FDE and the Denoising Diffusion Implicit Model (DDIM). Among them, the feature differential encoder SU-FDE is used to improve the accuracy of edge description and is combined with Siamese network contrastive learning to analyze the similarity and difference of image pairs. In addition, multi-scale information fusion using nested U2Net improves edge detail description. Identify and enhance key coarse features through the Spatial Attention (SA) \cite{Woo_2018_ECCV} module. Subsequently, the key feature maps are introduced into the encoder of the diffusion model to enhance the robustness to illumination and climate changes. DM stepwise sampling integrates key features effectively and iteratively generates more accurate CD maps. The main contributions of our work are summarized below:

\begin{itemize}
    \item [\textbullet]We propose SMDNet, a new model that combines Siamese encoder and DDIM architecture. It includes SU-FDE and denoising UNet, improving key feature detection, edge recognition, and overall model robustness.
    \item [\textbullet]The SU-FDE module of SMDNet uses the Siamese network and U2Net to extract multi-scale differential information from dual-temporal images, enhance the recognition of spatial correlation and difference, and improve edge detail detection capabilities.
    \item [\textbullet]SMDNet uses Denoising U-Net to learn pixel distribution under different lighting and weather conditions to improve model robustness and achieve good F1 scores of 90.99\%, 88.40\%, and 88.47\% on three public data sets.
\end{itemize}

The structure of the rest of the paper is as follows: Section II discusses related work in DL and DM applied to CD. Section III details the methodology we propose. Section IV covers the experimental setup and analysis of results, while Section V concludes the paper.

\section{Related Work}

\subsection{Deep Learning Based CD Methods}

With the swift advancement of deep learning technology in recent years, its strong potential in CD in RS images has received extensive research attention. In particular, the ability of CNN in pixel-level CD is outstanding in learning feature representations and identifying changed areas. In the realm of CD, the approaches can be categorized into two types: single-stream and dual-stream network methods.
    
\textbf{Single-stream network}: A singular deep learning network structure is employed to learn the evolving features between dual-temporal images. In this approach, a fully convolutional neural network (FCNN) \cite{long2015fully} primarily handles demanding prediction tasks. To achieve this, two dual-temporal images before and after changes are concatenated into a single input image, and the change detection feature map is extracted using a convolutional network.
Recognizing that the FCNN's sampling process may result in information loss and weak global perception ability, the method incorporates U-Net \cite{ronneberger2015u}. This involves merging dual-temporal images into a single image, fed into the network with modifications. A recurrent neural network is introduced in the skip connection to enhance responsiveness to temporal changes in perception ability.
PBSL \cite{shen2023pbsl} introduces a multimodal alignment approach to highlight relevant features and suppress irrelevant information.
To capture global and edge detail information, the UNet++ network \cite{zhou2018unet++}\cite{peng2019end} employs dense skip connections in the Encoder for change detection tasks, enhancing the segmentation network. The method adopts a deep supervision strategy to improve detection in high-resolution images. Nevertheless, it encounters challenges in understanding temporal data relationships, limiting its change detection capability.

\textbf{Dual-stream network}: The approach involves processing images from two temporal nodes through two parallel neural networks, establishing a coupled architecture. This empowers the network to discern features with significant material relationships, improving change detection (CD) map accuracy. Unlike a single network structure, the dual-stream network is divided into an asymmetric dual-stream network (pseudo-Siamese) and a Siamese network, each handling inputs from different time points. The asymmetric network learns distinct features with independent weights, while the Siamese network compares input similarities with shared weights adapted to different time series data analysis tasks.
Recurrent Convolutional Neural Networks (ReCNN) \cite{mou2018learning} integrate CNNs with Recurrent Neural Networks (RNNs) to form a pseudo-Siamese network, enabling the learning of spectral, spatial features, and temporal relationships. The Siamese network, in the dual-task constrained deep Siamese convolutional network (DTCDSCN) \cite{liu2020building}, utilizes Fully Convolutional Networks (FCNs) to extract multilevel features. Combining deep Siam FCNs with channel and spatial attention enhances the network's capability to capture unique features.
Addressing the lack of sufficient supervision for change feature learning, an intensely supervised Image Fusion Network (IFN) \cite{zhang2020deeply} is proposed using a pre-trained VGG16 as an Encoder through the Attention Module of the dual-stream architecture.
TCRL \cite{shen2023triplet} introduces a contrastive learning method to facilitate the interaction of global and local features, enhancing semantic saliency.
To enhance the integration of deep and shallow features, the densely connected Siamese network (SNUNet) \cite{fang2021snunet} merges Siamese network principles with UNet++. Introducing a novel parallel differential encoder, this network blends time-related information from bi-chronological images, exhibiting heightened sensitivity to subtle variances.

\subsection{Diffusion Model}
DDPM as a generative model, is focused on reconstructing data by inversely simulating the diffusion process of data from its true distribution to the noise distribution. During training, the model gradually masters the process of recovering from the noisy state to the original data. 
Compared with traditional deep learning frameworks such as single-stream or dual-stream networks, DDPM demonstrates multiple advantages: they can capture pixel changes under different time and meteorological conditions and enhance key features of the image when reconstructing the real data, which improves image contrast and clarity. 
DDPM has shown its superiority in synthesizing and recovering high-quality images. 
For example, 
In the field of CD in RS images, these models effectively distinguish real changes from pseudo-changes due to noise through an iterative denoising process, thus improving the detection accuracy of details and edges. Although there are not many cases of utilizing DM in the research literature on RS image change detection, recent innovative research has started investigating the application of DDPM in processing RS images, especially in feature extraction and construction of pre-trainer encoders for large-scale RS data. Inspired by the successful application of DDPM in other fields, this study proposes an innovative method for RS image change detection: the Siamese U2Net Denoising Diffusion Implicit Model(SMDNet).

\begin{figure*}[!t]
\begin{center}
    \includegraphics[width=17cm,height = 9cm]{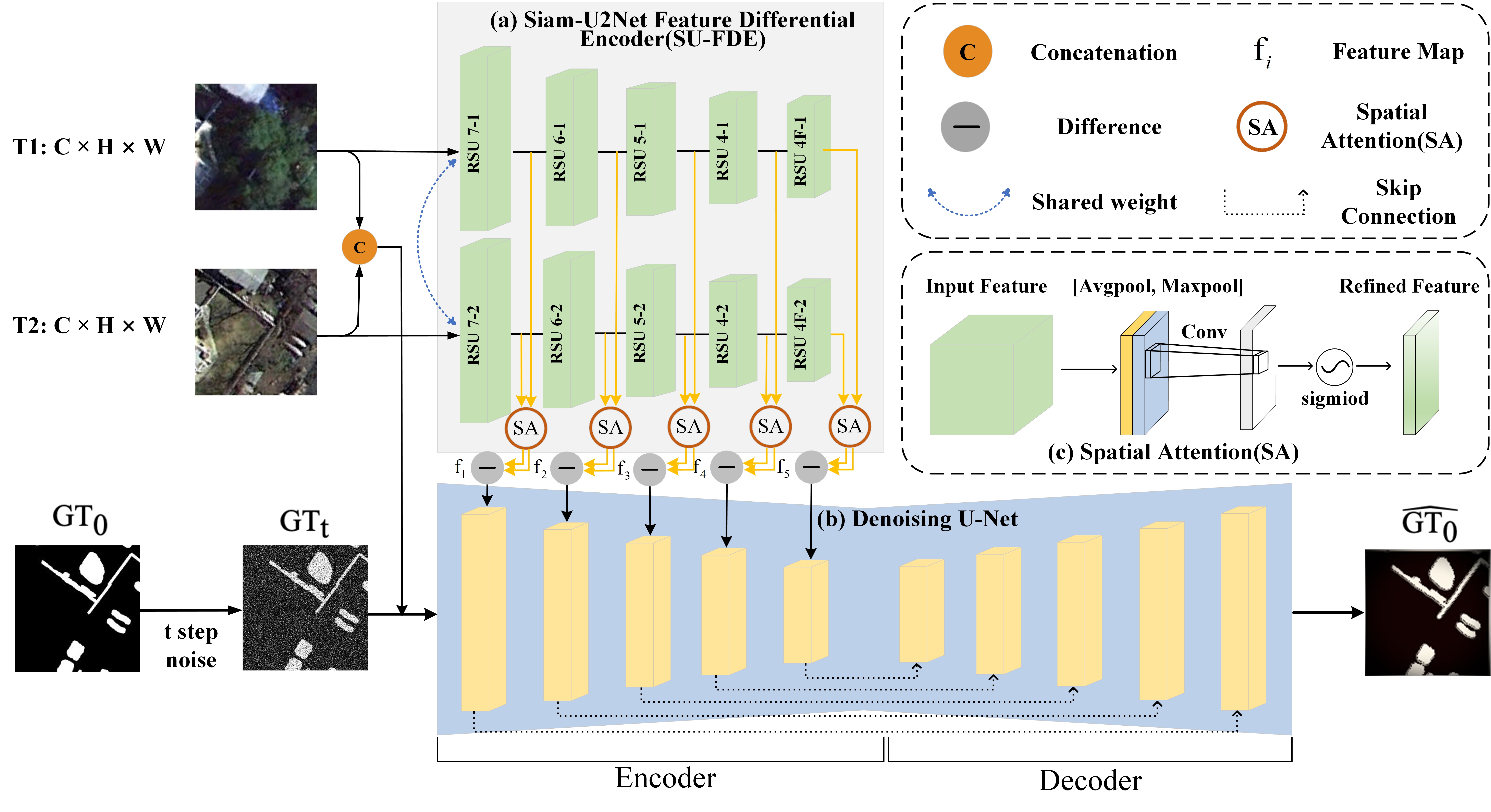}
\end{center}
\caption{The structural layout of the SMDNet network design. T1, T2, and $GT_{0}$ are the pre-change, post-change, and labeled images in the CDD dataset. (a) is the proposed Siam-U2Net Feature Differential Encoder (SU-FDE) for feature extraction of bitemporal image pairs; (b) is the Denoising UNet for denoising; (c) is the Spatial Attention Mechanism.}
\label{fig:SMDNet}
\end{figure*}

\section{Method}
This section will introduce the proposed SMDNet network. Initially, the overall architecture of the network is presented. The proposed feature differential encoder module (SU-FDE) is described in detail. Next is an explanation of the added attention mechanism. Finally, DDIM is briefly explained.
\subsection{Framework Overview}
Fig.\ref{fig:SMDNet} mainly comprises a SU-FDE and a denoising module (Denosing U-net). SU-FDE is a bitemporal U2-Net feature differential encoder. Distinct from conventional change detection methods and typical deep learning approaches is the process of letting the DM learn to denoise. After the attention mechanism emphasizes the critical features of the bitemporal image pairs, the differential calculation of the feature maps is performed to derive the difference map. The difference map and the noisy label map are fed into the diffusion model, which learns from the noise to produce clear detection results.

Due to the richer texture and geometric semantic information in high-resolution RS images compared to standard optical RS images, there is a heightened requirement for more advanced feature extraction capabilities. First, the bitemporal image pair (
T1, T2) is input into SU-FDE, resulting in a multi-scale feature map of the image pair. Spatial attention is then applied to reinforce key spatial features. Then $L_{1}$ calculates the distance between two feature maps with the same shape at different scales and obtains the difference map ($\widehat{f_{i}}$) of each layer. At the same time, the noise label map $GT_{t}$ is obtained by adding $t$ step noise to the binary classification (changed/unchanged) label $GT_{0}$. The bitemporal T1 and T2 and the added noise $GT_{t}$ are combined along the channel dimension before being inputted into the Encoder of Denoise-UNet (DU) to extract multi-scale features($I_{S})$. $(\widehat{f_{i}})$ and $(I_{S})$ have the same number of features. Add the features of the corresponding scales of $\widehat{f_{i}}$ and $I_{S}$ to obtain the fusion feature. Then, input the obtained fusion features into the decoder of DU to obtain the prediction result $\widehat{GT_0}\in \mathbb{R}^{C\ast W\ast H}$.
\begin{equation}
GT_0=DU(cat(T1,T2,x_t),t,\widehat{f_{i}})
\end{equation}
Here $cat(\cdotp)$ is the concatenation operator.

The change detection task is to detect changed and unchanged areas between pixels in an image. This is a discrete binary classification task, and in change detection tasks, changed areas usually look much smaller than unchanged areas. Therefore, simultaneously using Dice Loss and BCE Loss can improve the model's sensitivity to boundary pixels and overall pixel classification accuracy. The total loss here is:
\begin{equation}
L_{total} = {L}_{dice}(\widehat{{GT}_{0}},{GT}_{0})+{L}_{bce}(\widehat{{GT}_{0}},{GT}_{0})
\end{equation}
Here ${L}_{dice}$ is dice loss, ${L}_{bce}$ is BCE loss, $\widehat{{GT}_{0}}$ is the predicted label value, ${GT}_{0}$ is the ground true label value.

\subsection{Siam-U2Net Feature Differential Encoder}

\begin{figure}[!ht]
\begin{center}
    \includegraphics[width=8.5cm,height = 12.87cm]{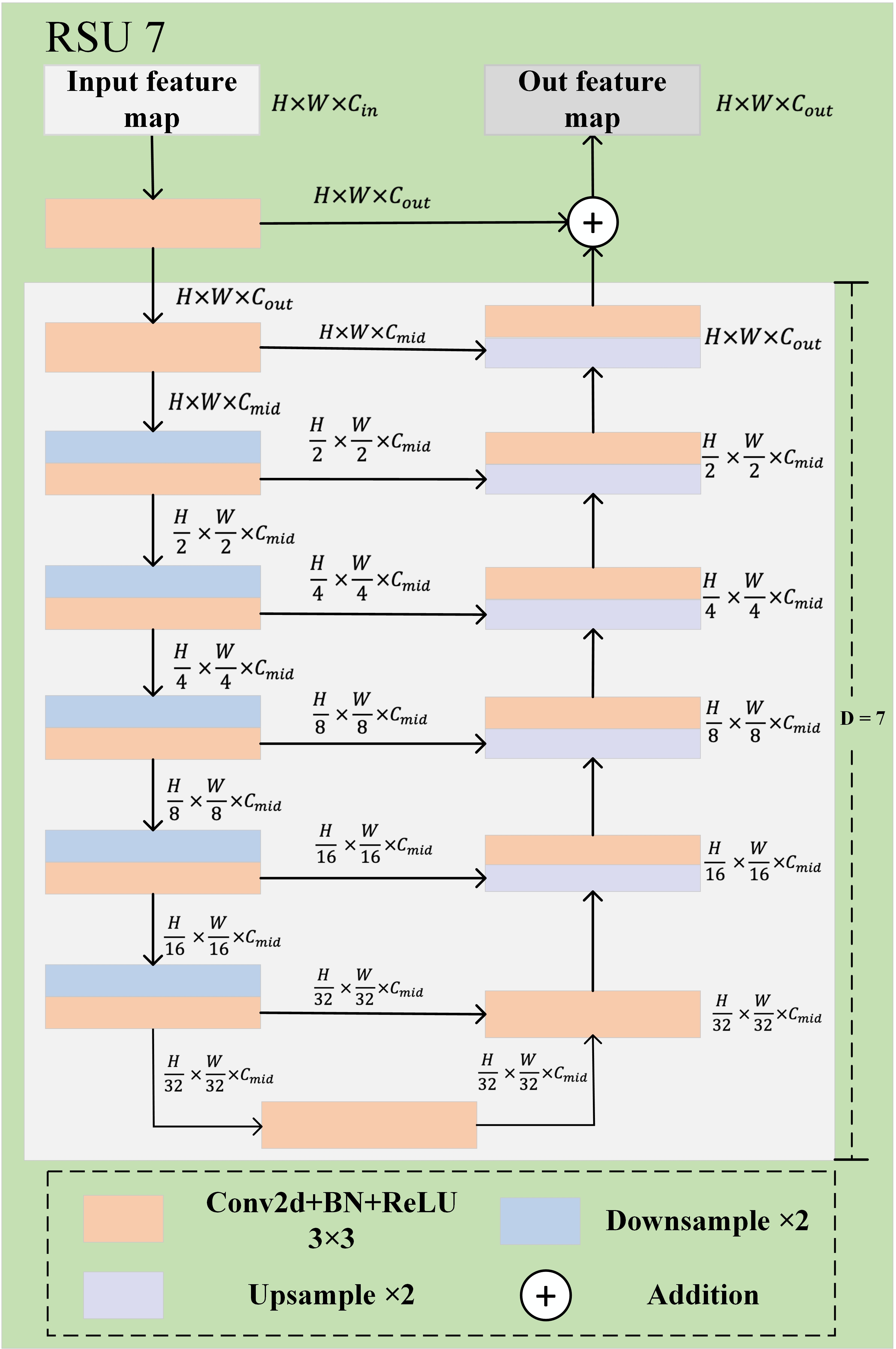}
\end{center}
\caption{Residual U Block (RSU). It adopts the encoder and decoder designed by U-Net, which is symmetrical and has D-layer depth. In the figure example, the depth is 7 layers for the RSU 7 block.}
\vspace{-.2cm}
\end{figure}

In the current RS image CD research domain, dealing with complex geographical environments and diverse ground object types (Such as roads, buildings, plants, lakes, etc.) is a crucial challenge. To overcome this challenge, researchers usually employ various advanced techniques, such as feature pyramid fusion, Inception modules, skip dense connections, and residual connections, to improve the performance of image detail capture and edge detection. In this context, we introduced the Siamese network and built a lightweight feature extractor named SU-FDE. This feature extractor aims to enhance the understanding of similarities and differences in images to improve the ability of multi-scale information fusion and edge detection. We achieved satisfactory results by fully pre-training SU-FDE and extracting the trained weights, which were subsequently used in the model's feature extractor and frozen.

The SU-FDE approach primarily consists of two main components:

\textbf{U-shaped structure feature extractor}: We design a feature encoder using a U-shaped structure containing 10 residual Residual U Block (RSU). RSU combines receptive fields of various sizes and can capture contextual information at multiple scales. These RSU blocks form a coherent U-shaped feature extraction framework, which helps extract feature information of complex geographical environments and diverse ground object types. These RSU blocks can be divided into two main structures: the first four layers and the fifth layer. The first four layers are 4 pairs of residual U blocks of different depths, namely RSU-7, RSU-6, RSU-5, and RSU-4. The numbers here (such as "7", "6", "5", "4") represent the depth (D) of the RSU block (as shown in Figure 2). This depth can be adjusted based on the resolution of the input feature map. In high-resolution remote sensing images, deeper RSU blocks can be selected to capture more detailed information. In addition, the fifth layer adopts another RSU-4F (right in Figure 2) structure, where “F” represents the dilated convolution version. These blocks are used to downsample to 16x and 32x deeper and to keep the resolution of the feature maps not reduced. We use dilated convolutions instead of traditional downsampling and upsampling. This helps prevent the loss of useful contextual information and ensures that the input and output resolutions of the RSU-4F remain consistent.

\textbf{Multi-level feature difference and fusion}: In the second part, we leverage a spatial attention module (SA) to augment the sensitivity of the feature map in each layer of the Siamese network towards spatial changes in the context of the CD task. The feature map is spatially enhanced through the SA module, which focuses on critical areas for CD in RS images. The SA module combines the information gathered through maximum and average pooling to generate a salient feature map. This salient feature map is then processed by the sigmoid activation function, resulting in an attention map of the same size as the original feature map. Attention maps highlight important spatial regions and suppress less relevant areas, improving the model's ability to recognize subtle changes. We measure the difference between processed feature maps through the L1 norm and effectively fuse the difference features into the encoding layer of Denoise-UNet, thereby enhancing change detection performance while maintaining computational efficiency.

Combining the Siamese network and U2Net, the SU-FDE module is more potent for detecting RS image change. It improves the model's comprehension of image similarities and differences while enhancing the fusion of multi-scale information and edge detection capability.

\subsection{Denoising Diffusion Implicit Models}

DDIM serves as another crucial component within the SMDNet model. It plays a role in image denoising and enhancing the model's resilience to lighting and meteorological conditions changes. DDIM proposes an accelerated diffusion model based on DDPM. This generative model learns the noise distribution and removes noise in the image, improving change detection accuracy. Here is a brief overview of DDPM and DDIM.

DDPM consists of a forward Markov denoising process, denoted as $q$, and a reverse denoising process, denoted as $p$ (\cite{ho2020denoising}). The forward process starts from $x_0$ with a label map $q(x_0)$ at $T$ time steps, gradually introducing noise using a Gaussian distribution:
\begin{equation}
    q({x}_{1}:T|{x}_{0}) = \displaystyle\prod_{t=1}^{T}q({x}_{t}|{x}_{t-1})
\end{equation}
where:\begin{equation}
    q({x}_{t}|{x}_{t-1}) = N({x}_{t};\sqrt{1-\beta _t}{x}_{t-1},\beta _tI)
\end{equation}
Here,  $x_0$ is the label map. $x_t$ represents the state of the image at time step $t$, and $\beta_t$ is a noise scale parameter.

In Equation 3, as pointed out by \cite{ho2020denoising}, it is stated that $x_t$ can be sampled for any time step $t$, eliminating the need to reuse $q$. Here ${\alpha }_{t}:=1-{\beta }_{t}$, and ($\bar{{\alpha }_{t}}:\textstyle\prod_{s=0}^{t}{\alpha }_{t}$), can be written:
\begin{equation}
    q({x}_{t}|{x}_{0})=N(x_t;\sqrt{{\alpha }_{t}}x_0,(1-\bar{{\alpha}_{t}})I)
\end{equation}
\begin{equation}
    x_t=\sqrt{{\alpha }_{t}}x_0+\sqrt{(1-\bar{{\alpha }_{t}})}\epsilon,\epsilon \in N(0,I)
\end{equation}
$\alpha_t$ is defined as $1 - \beta_t$, where $\alpha_t$ is the complementary part of the noise scale parameter. DDIM modulates the label map and the impact of the noise introduced at each time step. $\bar{\alpha}_t$ is the cumulative product of $\alpha_t$ from 0 to $t$. That is the total amount of noise added at time step $t$. $\epsilon$ is a random noise added to the image.

In simpler terms, the forward process of DDPM gradually introduces noise to the image, transforming it from its initial state to a fully noisy state over $T$ time steps.

The reverse process of DDPM involves the neural network's prediction of the noise added during the forward process, enabling the step-by-step reconstruction of the original image from its noisy state. This process portrays the neural network's ability to identify and subtract noise at each time step efficiently. The reverse process, utilizing Bayes' theorem, discovers $p(x_{0:T})$ and decomposes it according to the chain rule formula:
\begin{equation}
    p(x_{0:T}) = p(x_T)\displaystyle\prod_{t = T}^{1}p({x}_{t-1}|x_t)
\end{equation}
where ($p(x_T)\in N(0,I)$ ) is a standard normal distribution.
\begin{equation}
p(x_{t-1}|x_T)=N(x_{t-1};\mu (x_t,t),\sum (x_t,t))
\end{equation}
$\mu(x_t, t)$ and $\sum(x_t, t)$ represent the mean and variance used in the reverse process, respectively.

Marginalize through $p(x_{0:T})$ to get the marginal probability $p(x_0)$ of $x_0$:
\begin{equation}
    p(x_0)=\int p(x_{0:T})dx_{1:T}
\end{equation}
By applying Jensen's inequality, we derive an evidence lower bound (ELBO) for the logarithm of the likelihood function. We then train the backward process to align its distribution with the forward process distribution: $-L(x_0) <= log(p(x_0))$:

\begin{equation}
\begin{aligned}
L(x_0) &= \text{Eq}[L_{T}(x_0) + \\
&\quad \displaystyle\sum_{t>1} D_{KL}(q(x_{t-1}|x_t, x_0) || p(x_{t-1}|x_t)) - \log p(x_0|x_1)]
\end{aligned}
\end{equation}

Where\begin{equation}
    L_{T}(x_0) = D_{KL}(q(x_{T}|x_0)||p(x_{T}))
\end{equation}

There exist multiple approaches to parameterize $\mu (x_t,t)$(Eq.(8)) the priors, and we can predict the formula with a neural network$\mu (x_t,t)$, which can also predict noise$\epsilon $. Here, we directly expect $x_0$ instead of noise$\epsilon $ and calculate the $\mu (x_t,t)$. In \cite{ho2020denoising}, the training objective of optimizing network parameters is simplified, and the optimization objective is proposed; additionally, a reweighted loss function is introduced:

\begin{equation}
    E_{t,x_0,\epsilon }\parallel\epsilon -\epsilon_{x_t,t}\parallel^2
\end{equation}

After the diffusion model is trained, $x_{T}$ is sampled from $N(0,I)$, and $x_t$ is denoised and iterated to obtain a new $x_{0}$ at each time step t:
\begin{equation}
\begin{aligned}
    x_{t-1}&=\sqrt{\bar{\alpha _{t-1}}}(\frac{x_t-\sqrt{(1-\bar{\alpha _t})}\epsilon _{(x_t,t)}}{\sqrt{\bar{\alpha _t}}})+ \\
    &\quad \sqrt{1-\bar{\alpha }_{t-1}-\sigma^2_ {t}}\epsilon _{(x_t,t)}+\sigma _t\epsilon
\end{aligned}
\end{equation}

In \cite{ho2020denoising}, DDIM uses variance $\sigma^2$ as a hyperparameter that can be manually adjusted, and different effects can be obtained by adjusting $\sigma^2$.

\begin{equation}
    \sigma ^2_t=\frac{(1-\alpha _t)(1-\bar{\alpha }_{t-1})}{1-\bar{\alpha }_{t}}
\end{equation}

In DDPM:
\begin{equation}
    \sigma_t=
\sqrt{(1-\bar{\alpha }_{t-1})/(1-\bar{\alpha }_{t})}\sqrt{(1-\bar{\alpha }_t)/\bar{\alpha }_{t-1}}
\end{equation}
In ddim, setting $\sigma_t = 0$ becomes deterministic sampling, and the generation process is deterministic. When $x_{t-1}$ and $x_0$ are given, the forward process becomes deterministic, so the resulting model is an implicit probability model. Samples are generated from latent variables using a fixed process (from $x_T$ to $x_0$). When sampling When the length is much smaller than $T$, the computational efficiency is substantially enhanced as a result of the iterative nature of the sampling process.

As one of the key components of the SMDNet model, leveraging DDIM accelerates the noise removal process in remote sensing images, consequently enhancing the overall performance of change detection. DDIM gradually reduces noise through multiple iterative diffusion processes, thereby removing noise and improving the image. This capability contributes to the model's adeptness in detecting subtle changes in RS images, thereby enhancing the accuracy and reliability of the detection results.
\section{Experimental setup and Result analysis}
\subsection{Datasets and comparisons}
In our experiments, we used three popular public datasets to evaluate the performance of our model across diverse lighting conditions, seasonal changes, different resolutions, and changing terrains and scenes. They are the LEVIR-CD dataset, the DSIFN-CD dataset, and the CDD dataset. For the consistency of the test, the data set was pre-processed, cropped into image pairs of 256x256 size, and randomly divided into three parts: train/test/val with 0.75/0.2/0.05.

\textbf{LEVIR-CD Dataset}\cite{rs12101662} collected 637 pairs of image patches with very high resolution from Google Earth (GE). Each patch had a pixel resolution of 0.5m and a size of 1024x1024 pixels. These image pairs come from more than 20 regions in the United States, with time blocks ranging from 5 to 14 years, with significant land changes and building growth, migration, etc. Due to its extensive period and diverse range of building types, this dataset presents a significant challenge for large-scale remote sensing building change detection.

\textbf{DSIFN-CD Dataset}\cite{ZHANG2020183}is six large bitemporal high-resolution images collected from Google Earth. It covers six cities in China (i.e., Beijing, Chengdu, Shenzhen, Chongqing, Wuhan, and Xi'an) that contain significant differences among the land objects collected. The five large images in the dataset were segmented into 394 sub-image pairs, each measuring 512x512 in size. Following data augmentation, a total of 3940 bitemporal image pairs were generated. Among them, the image pairs of Xi'an are cropped to 48 as the test set. This dataset is challenging due to its abundance of land objects and images that vary seasonally.

\textbf{Google Earth Change detection dataset(CDD)} \cite{lebedev2018} has remote sensing imagery of seasonal changes in an area. The dataset comprises 11 pairs of images, including 7 image pairs with dimensions of 4725x2700 pixels and 4 pairs of 1900x1000 pixels. The final output is a set of 16,000 cropped images, each measuring 256x256 pixels. it is divided into 10,000 image pairs for the training set and 3,000 for testing and validation. With a high spatial resolution ranging from 3 to 100 cm, it offers detailed insights into changes in everyday structural objects like cars, buildings, roads, and seasonal variations in natural objects, from individual trees to extensive forest areas.

\begin{table}
\caption{dataset statics}
\label{tab:dataset_detail}
\begin{center}
\begin{tabular}{cccccc}
\hline
Dataset & Year & Resolution & Image Size & Pairs \\ \hline
LEVIR-CD\cite{rs12101662}& 2020 &  0.5m  &  1024$\times$1024  &   10192    \\
DSIFN-CD\cite{ZHANG2020183}& 2020 &   2m   &  512$\times$512  &15760     \\
CDD\cite{lebedev2018} & 2018 & 0.03 $\sim$ 1m & 4725$\times$2200 &16000 \\ \hline
\vspace{-.2cm}
\end{tabular}
\end{center}
\end{table}

To validate the reliability of our method, we conducted comparisons with eight state-of-the-art CD methods from recent years on the same three public datasets, using image block sizes consistent with those in the referenced papers. These methods include Fully Convolutional Early Fusion (FC-EF), Fully Convolutional Siamese-Difference (FC-Siam-diff), Fully Convolutional Siamese-Concatenation (FC-Siam-Conc), Spatial-Temporal Attention Neural Network (STANet), A Deeply Supervised Image Fusion Network (IFNet), Siamese NestedUNet (SNUNet), BIT, and Changeformer. Each method demonstrates effectiveness in various aspects of CD through different network architectures. FC-EF, FC-Siam-diff, and FC-Siam-Conc are three pioneering CD methods based on deep learning, all proposed in the same research paper. STANet achieves better spatiotemporal relationship modeling by introducing multi-scale sub-region division and obtaining long-range spatiotemporal information in the self-attention module. IFNet employs a dual-stream architecture to extract deep features and then fuses these deep features with differential features via an attention module, reconstructing the CD map in the process. SNUNet, which merges the Siamese Network with NestedUnet, enhances CD performance by integrating deep and shallow features. This integration is achieved through efficient information transmission and the application of attention modules, leading to a more effective feature synthesis for CD tasks. A simple CNN backbone (ResNet18) is combined with an end-to-end transformer to enhance the CD recognition of changing areas of interest. In addition, another Siamese network combines the Changeformer with a hierarchical Transformer encoder and MLP decoder to achieve efficient, multi-scale CD, effectively capturing the detailed, long-range information required for precise CD tasks.

\begin{table*}
    \centering
    \caption{Quantitative comparison on three datasets. The best-performing values are highlighted in bold. All scores are presented in percentage format (\%).}
    \resizebox{1\textwidth}{!}
    {
    \begin{tabular}{cccccccccccccccc}
    \toprule
    \multicolumn{1}{c}{} &
    \multicolumn{5}{c}{\textbf{LEVIR-CD}}  &  
    \multicolumn{5}{c}{\textbf{DSIFN-CD}} &
    \multicolumn{5}{c}{\textbf{CDD}}\\
    \cmidrule(lr){2-6} \cmidrule(lr){7-11} \cmidrule(lr){12-16}
    & Pre. & Rec. & F1 & IoU & OA &  Pre. &  Rec. & F1 & IoU & OA & Pre. &  Rec. & F1 & IoU & OA \\ 
    \midrule
    FC-EF \cite{Daudt2018} & 86.91 & 80.17 & 83.40 & 71.53 & 98.39  &  72.61 & 52.73 & 61.09 & 43.98 & 88.59 & 80.81 & 64.39 & 71.67 & 55.85 & 85.85   \\
    FC-Siam-diff \cite{Daudt2018} & 89.53 & 83.31 &  86.31 & 75.92 & 98.67  &  59.67 & 65.71 & 62.54 & 45.50 & 86.63 & 85.44 & 63.28 & 72.71 & 57.12 & 87.27\\
    FC-Siam-Conc \cite{Daudt2018} & 91.99  & 76.77  & 83.69 & 71.96 & 98.49  & 66.45 & 54.21 & 59.71 & 42.56 & 87.57 & 82.07 & 64.73 & 72.38 & 56.71 & 84.56 \\
    STANet \cite{rs12101662} & 83.81  &  \textbf{91.00}  &  87.26 & 77.40 &  98.66 & 67.71 & 61.68 & 64.56 & 47.66 & 88.49 & 89.37 & 65.02 & 75.27 & 60.35 & 82.58 \\
    IFNet \cite{ZHANG2020183} &  \textbf{94.02} & 82.93 & 88.13 & 78.77 & 98.87 & 67.86 & 53.94 & 60.10 & 42.96 & 87.83 &   -  &  -  &  -  &  -  &  -   \\
    SNUNet \cite{fang2021snunet}  & 89.18 & 87.17 & 88.16 & 78.83 & 98.82 & 60.60 & 72.89 & 66.18 & 49.45 & 87.34 &  -  &  -  &  -  &  -  &  -  \\
    BIT \cite{Chen2022} & 89.24 & 89.37 & 89.31 & 80.68 & 98.92  &  68.36 & 70.18 & 69.26 & 52.97 & 89.41 & \textbf{92.04} & 72.03 & 80.82 & 67.81 & 96.59 \\ 
    ChangeFormer \cite{bandara2022transformer} &  92.05 & 88.80 & 90.40 & 82.48 & 99.04 & 88.48 & 84.94 & 86.67 & 76.48 & \textbf{95.56} &  -  &  -  &  -  &  -  &  -   \\
    \midrule
    SMDNet(Ours) & 92.71 & 85.89 & \textbf{90.99} & \textbf{82.71} & \textbf{99.17}
& \textbf{88.51} & \textbf{88.46} & \textbf{88.40} & \textbf{80.91} & \textbf{96.56} & 89.13 & \textbf{87.35} & \textbf{88.47} & \textbf{83.30} & \textbf{99.29}
\\ 
   \bottomrule
    \end{tabular}
    }
    \label{tab:comparison_sotas}
\end{table*}

\subsection{Evaluation Metrics} 
In our evaluation framework, the primary metric for assessing performance is the F1-score derived from the test's precision and recall values. The formula to calculate the F1-score(F1) is:

\begin{align}
F1-score = 2\times \frac{Precision\times Recall}{Precision + Recall}
\end{align}

Furthermore, we also provide additional metrics for a comprehensive assessment, and These metrics take into account true positives (TP), false positives (FP), false negatives (FN), and true negatives (TN) in their calculations. TP represents correctly identified changes, FP indicates incorrectly marked changes, FN denotes missed actual changes, and TN refers to correctly identified non-changes. Specifically included are Precision(Pre.) and Recall(Rec.) for the change category, the Intersection over Union (IoU) for the same, and the overall accuracy (OA). The formulas for these evaluation metrics are as follows:

\begin{align}
Precision &= \frac{TP}{TP + FP}\\
Recall &= \frac{TP}{TP+FN}\\
IoU &= \frac{TP}{TP+FN+FP}\\
OA &= \frac{TP+TN}{TP+TN+FN+FP}
\end{align}

\subsection{Experimental details and analysis}
Our network SMDNet is implemented using the PyTorch framework and executed on a single NVIDIA A100 GPU. We configured the batch size to 8 during training to ensure memory usage. The AdamW optimizer is used with the learning rate set to 1e-4, and the corresponding weight attenuation factor is also set to 1e-4 to prevent overfitting. The warm-up period is set to 1\% of the total period, and the learning rate is updated according to the cosine annealing schedule. The model is regularly verified every 10 epochs. During training and testing, DDIM selects 10 key points from 1000 time steps to guide the denoising process.

\subsubsection{LEVIR-CD Dataset Analysis}
LEVIR-CD is a dataset of high-resolution images with an extended period, rich in building types and changing scenes. In experiments conducted on the LEVIR-CD data set, as shown in Table 2, the SMDNet model achieved 92.71\% and 90.99\% in Pre. and F1, indicating that the model performed well in detecting subtle significant architectural changes, such as The emergence of new buildings or the demolition of old buildings. However, the model performed slightly worse regarding recall (85.89\%) and intersection-over-union (IoU, 82.71\%). This may be because there are fewer samples of the changed category in the dataset, causing the model to prefer predicting the "no change" category. It improves accuracy but reduces recognition of actual changes. Here, IFNet leads in Pre. with 94.02\%, but its performance in Rec. and IoU (82.93\% and 78.77\%, respectively) is slightly insufficient, which reflects the performance trade-off between different models. STANet performs outstandingly regarding Rec. (91\%),  indicating that the model may need to be selected and optimized according to specific application scenarios in different change detection tasks.
Fig. \ref{fig: LERIVandDSIFNduibi}(a, b, and d) shows the images before and after the change and the ground truth, respectively. Fig. \ref{fig: LERIVandDSIFNduibi} (a) has the initial surface state, such as lakes, farmland vegetation, and buildings. In (b), changes that may involve building expansions, land use conversions, or seasonal changes can be observed. The ground truth (d) marks the areas of change and provides a benchmark for evaluating the model’s performance in different seasonal scenarios. It can be seen from the Visualization effect in Fig. \ref{fig: LERIVandDSIFNduibi} (c) that the SMDNet model can accurately identify the actual change areas in most cases, consistent with the real change annotations on the ground. The model's accuracy is critical in detecting new construction or demolition of buildings and is essential for monitoring urban expansion or redevelopment activity. However, the imbalance of classes in the dataset may cause the model to be too conservative when dealing with broad unchanged areas, lacking sensitivity to subtle changes and hiding potential risks of missed detections. Therefore, future research should explore data augmentation or resampling techniques to balance class distribution or develop more sensitive change detection algorithms to improve recall and IoU to ensure the model maintains high accuracy without sacrificing detection capabilities.

\begin{table*}
    \centering
    \caption{Model depth results at different layers on DSIFN-CD and CDD datasets. All scores are presented in percentage format (\%).}
    \resizebox{1\textwidth}{!}{
    \begin{tabular}{cccccccccc} 
    \toprule
    \multirow{2}{*}{\textbf{Layer Depth}} & 
    \multicolumn{4}{c}{\textbf{DSIFN-CD}} &
    \multicolumn{4}{c}{\textbf{CDD}} &
    \multirow{2}{*}{\textbf{Parameters}} \\
    \cmidrule(r){2-5} \cmidrule(l){6-9}
     & Precision & Recall & F1-score & OA & Precision & Recall & F1-score & OA &   \\ 
    \midrule
    4-layer & 86.27 & 86.27 & 85.35 & 95.90  &  82.51 & 81.84 & 81.81 & 99.00 &  12.47M \\
    5-layer & 88.46 & 88.22 & 87.40 & 96.56  &  85.07 & 85.03 & 84.64 & 99.18 & 20.12M \\
    6-layer & 91.18 & 89.48 & 89.05 & 97.67  &  89.88 & 89.26 & 89.20 & 99.43 &  29.69M \\
    \bottomrule
    \end{tabular}
    }
    \label{tab: Layer Depth Results}
\end{table*}

\subsubsection{DSIFN-CD Dataset Analysis}
The DSIFN-CD dataset has ground object scenes that change with seasons in different cities. As can be seen from Table 2, SMDNet surpasses other comparison methods on multiple key performance indicators. Especially in terms of precision and recall, SMDNet achieves a higher balance, which shows that it can maintain a high change detection rate while reducing false detections. SMDNet demonstrates its excellent overall performance in terms of F1 score, which is an essential indicator for evaluating the accuracy of change detection methods. On the IoU metric, combined with its high precision and recall, it performs well in distinguishing changing areas from non-changing areas. As seen from the visualization in Fig. \ref{fig: LERIVandDSIFNduibi}, the overall performance of our method in change detection is positive. The model can more accurately identify the changing areas between images, and the predicted shape is consistent with the actual change boundaries of ground objects, showing the model's good sensitivity to the edges of ground objects. However, there are also certain shortcomings. For example, some misjudgment areas may appear in some complex backgrounds. These areas may be caused by the model's sensitivity to subtle changes in the image. In addition, it can be seen from Figure 4 that the detailed part of the prediction may capture tiny changes in labels that are not annotated, which may be because the model has a certain sensitivity to these small-scale changes when extracting deep features. At the same time, it has specific suppression on some shadow changes.

\subsubsection{CDD Dataset Analysis}
The CDD dataset is mainly a seasonal variation dataset developed for RS. Observation of the graph shows that seasonal changes and different lighting conditions may cause spurious changes, thus challenging the model's ability to discern real changes. Prediction result (c) shows the sensitivity of our model SMDNet to various changes. In the first and second columns, our model shows high sensitivity to changes in snow-covered vegetation areas and building boundaries and its ability to resist spurious changes in shadowing.  In the third column (c) of Fig. \ref{fig: LERIVandDSIFNduibi}, the change in the edge of the water body is successfully identified. The figure shows that the model SMDNet can predict the boundary changes of objects under different weather conditions and shows good sensitivity and accuracy.
The performance evaluation results of the model in Table 2 show its effectiveness on the CDD dataset. Although the Pre. (89.13\%) It lags slightly behind the BIT method, reflecting the model's efficiency in noise suppression and classification accuracy. At the same time, SDMNet also performs well regarding Rec., reaching 87.35\%, ensuring that most real changes are successfully detected and avoiding missed detections. The F1 is 88.47\%, balancing precision and recall, making it a reliable indicator of the model's overall performance. The intersection-over-union (IoU) ratio is 83.3\%, indicating that the model has high consistency between predicted and actual change areas. Furthermore, the overall accuracy (OA) reached 99.3\%, demonstrating the model's high prediction accuracy on the entire dataset. However, OA may be overestimated due to the prevalence of unchanged regions in the dataset, highlighting the importance of comprehensive and accurate model evaluation using multiple performance metrics.

\subsection{Ablation experiments and analysis}

This subsection will help better understand the effectiveness of the proposed architectural and modular changes on the performance of SMDNet models in remote sensing image change detection tasks. We designed experiments from three aspects: 1. The impact of different layer depths in SU-FDE and Denoising U-Net; 2. Mainly the impact of adding different attention mechanisms to the model; 3. Performance evaluation of added components in the model; 4. The impact of the choice of diffusion step size in the diffusion model on feature extraction in the data set. The ablation experiments used four evaluation metrics: F1-score, Precision, Recall, and OA.

\subsubsection{Effect of layer depth}

The effect of different layers (four, five, and six) of the SU-FDE and Denoising U-Net on the performance of remote sensing image change detection in the SMDNet model was first evaluated. The number of channels for the Three types of depths were configured as follows: four-layer network (64, 64, 64, 128), five-layer network (64, 64, 64, 128, 128), and six-layer network (64, 64, 64, 128, 128, 256). These experiments were conducted on the CDD and DSIFN-CD datasets to explore the relationship between network depth and model performance and the number of parameters.

The results of the two datasets, DSIFN-CD and CDD in Table \ref{tab: Layer Depth Results} show that the F1 score and accuracy of the model increase as the number of layers increases, and the number of parameters also increases accordingly. Regarding the evaluation metrics, the 6-layer SMDNet performs well on the DSIFN-CD and CDD datasets. The complexity of the scene is higher due to the composite nature of the dataset, which includes a variety of ground objects such as buildings, roads, forests, and lakes. The 6-layer configuration demonstrates that the deeper model structure can extract more complex and abstract features, resulting in good evaluation metrics. However, it is interesting to note that the number of parameters in the 6-layer network is about 29.69M, which is about 2.4 times more than the 4-layer network (approximately 12.47M). The 5-layer network is not as good as the 6-layer network regarding F1 score and accuracy, but it requires fewer parameters, about 20.12M. Balancing performance and parameter efficiency, the 5-layer network structure maintains a certain level of performance while effectively controlling the model size to ensure the practicability of practical applications.

\begin{table}
\caption{The impact of using different attention mechanisms in SU-FDE on SMDNet on the CDD dataset}
\label{tab:CDD dataset about Attension}
\setlength{\extrarowheight}{1pt}
\begin{center}
\resizebox{\columnwidth}{!}{
\begin{tabular}{ccccc}\hline
            & F1-score & Precision & Recall & OA  \\ \hline
SU-FDE(+NL) & 86.20 &  86.31  &  86.78  &  99.24    \\
SU-FDE(+AX) & 86.33 & 86.30 & 87.14 & 99.27  \\
SU-FDE(+ECA) & 86.55 & 86.14 & 87.85 & 99.27   \\
Ours & 88.47 &  89.13  &  87.35  & 99.29\\ \hline
\end{tabular}
}
\end{center}
\end{table}

\subsubsection{The impact of attention mechanism on the model}

\begin{figure}[t]
    \centering
    \includegraphics[width=8.5cm,height = 6cm]{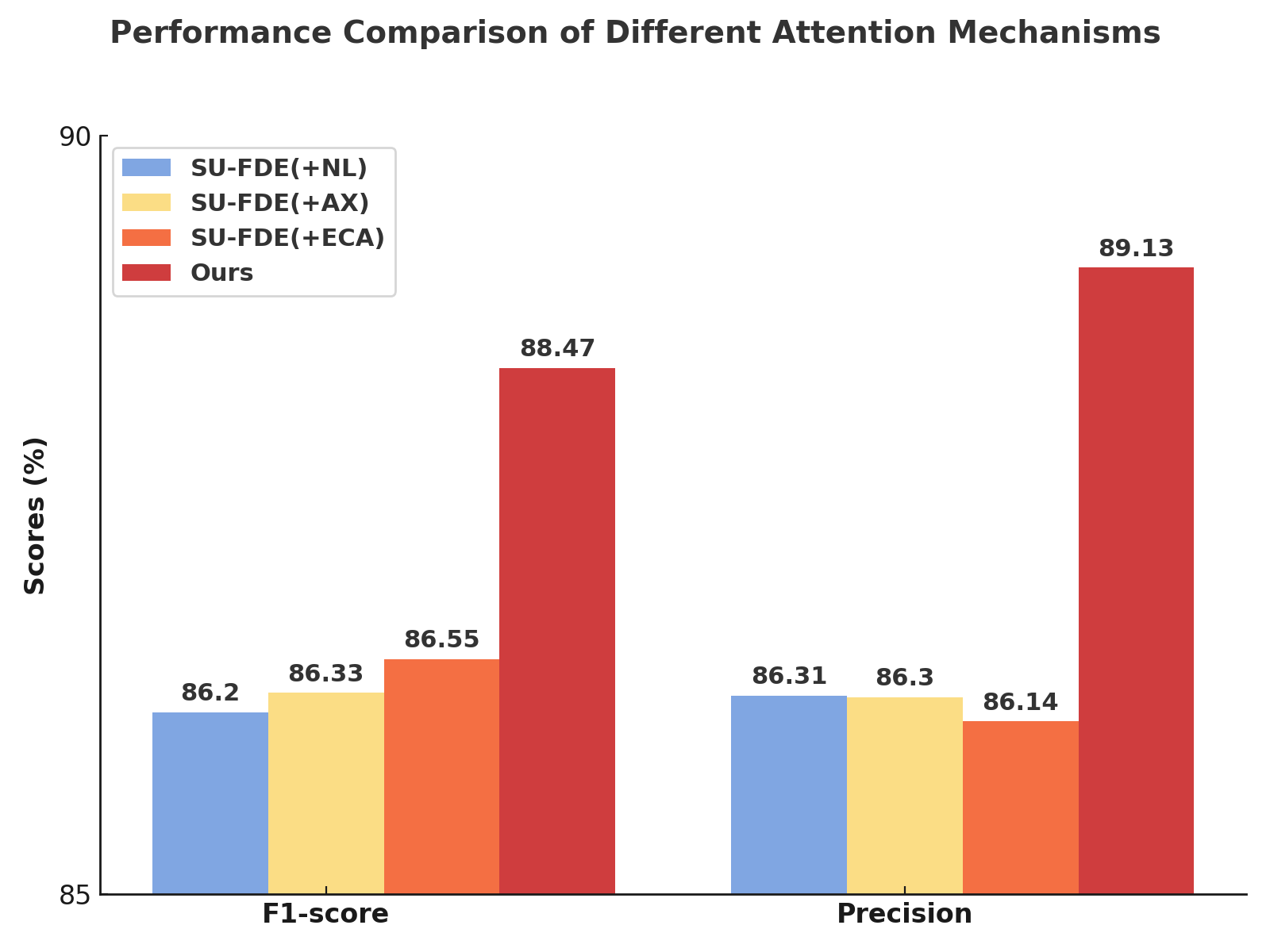}
\caption{Histogram comparison on CDD dataset using different attention mechanisms in SU-FDE of SMDNet}
\vspace{-.2cm}
\label{fig:bar atten}
\end{figure}

\begin{figure*}[t]
    \centering
    \includegraphics[width=18cm,height = 6.87cm]{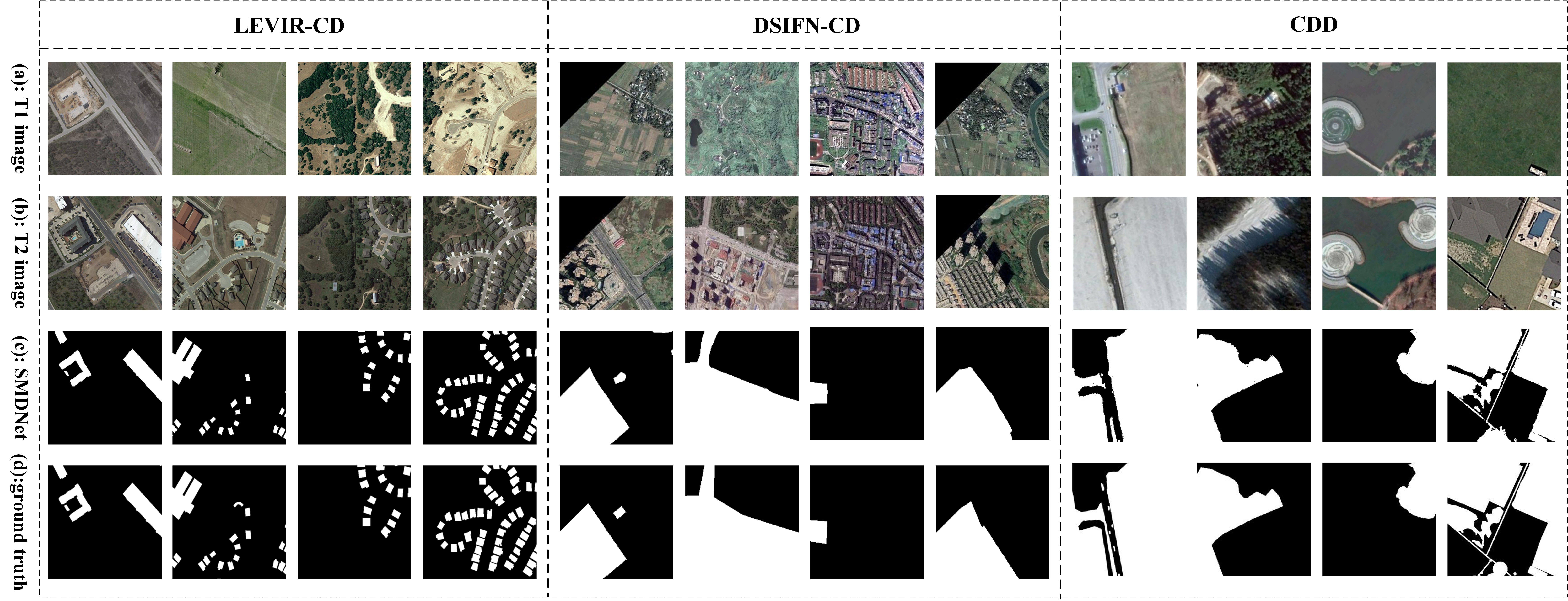}
\caption{Visualization of experimental results on three datasets LEVIR-CD, DSIFN-CD, and CDD.}
\vspace{-.2cm}
\label{fig: LERIVandDSIFNduibi}
\end{figure*}

We conduct ablation experiments using different attention mechanisms after each Feature Differential Extractor (SU-FDE) layer and evaluate their impact on model performance on the CDD dataset. As shown in Table IV and Fig. \ref{fig:bar atten}, we examine Non-Local (NL)\cite{wang2018nonlocal}, Axial Attention (AX)\cite{ho2019axial}, Efficient Channel Attention (ECA)\cite{wang2020eca}, and Spatial Attention (SA) mechanisms. NL enhances the network's ability to integrate long-range features by capturing global dependencies; AX attention maintains computational efficiency when processing large-scale data; the ECA mechanism maintains a low parameter burden while enhancing channel correlation; and the SA mechanism maintains a low parameter burden in space This enhances the recognition of local details and boundaries of the image. Experimental results show that adding any attention mechanism can improve the F1-score, Precision, Recall, and OA of SU-FDE. In particular, when the spatial attention mechanism (SU-FDE+(SA)) is integrated, the model performance is most significantly improved, with F1-score increasing to 88.47\%, Precision increasing to 89.13\%, and Recall reaching 87.35\%. At the same time, OA also Improved to 99.29\%. This result highlights the detailed sensitivity of the spatial attention mechanism in capturing spatial dependencies in images to enhance feature extraction capabilities and help maintain the spatial consistency of detected changes, which helps interpret and understand changes. The spatial attention mechanism can potentially improve model performance compared with other attention mechanisms.

\begin{table}[t]
\caption{Performance comparison of different components on the CDD dataset}
\label{tab:CDD dataset about Moudle1}
\setlength{\extrarowheight}{4pt}
\begin{tabularx}{8.9cm}{>{\centering\arraybackslash}m{1.3cm}
  >{\centering\arraybackslash}m{1cm}
  >{\centering\arraybackslash}m{0.4cm}
  >{\centering\arraybackslash}m{0.4cm}
  >{\centering\arraybackslash}X
  >{\centering\arraybackslash}X
  >{\centering\arraybackslash}X
  >{\centering\arraybackslash}X}\hline
Ablation Settings & SU-FDE & SA & DU & F1 & Pre. & Rec. & OA  \\ \hline
1 & \ding{51} & \ding{55} & \ding{55} & 71.77 & 67.72 & 82.96 & 98.68 \\
2 & \ding{55} & \ding{55} & \ding{51} & 80.97 & 81.39 & 80.41 & 98.77 \\
3 & \ding{51} & \ding{55} & \ding{51} & 84.64 & 85.07 & 85.04 & 99.19 \\
Ours & \ding{51} & \ding{51} & \ding{51} & 88.47 & 89.13 & 87.35 & 99.29 \\ \hline
\end{tabularx}
\end{table}

\subsubsection{Module ablation analysis}

Table V presents the results of our ablation experiments on different model components to evaluate their impact on the final model performance.
The first row contains only models of the SU-FDE encoding component designed to capture key features in the image. We equipped a simple decoder to form a complete network to test its performance. This component model achieved 71.77\%, 67.72\%, 82.96\%, and 98.68\% in F1-score, Precision, Recall, and OA, respectively. These results illustrate the potential of the SU-FDE encoder in extracting effective features. In the second row, we adopt the Denoising U-Net (DU) in the diffusion model as our main architecture. Compared to using only the SU-FDE encoder, this model significantly improves F1 score and accuracy. At the same time, the recall decreases slightly, which may indicate that the DU component is better at suppressing noise and improving the model's discriminative ability. The model in the third row combines the SU-FDE encoding component and the denoising U-Net (DU), improving all performance metrics. SU-FDE improves model accuracy by introducing deep feature details into the diffusion model. Finally, our proposed SMDNet (Ours) method integrates all components, where adding the SA mechanism improves detection efficiency by prioritizing spatial information when processing data. The last row shows that the comprehensive model performs optimally on all performance indicators. Compared with only DU, the F1 score increased by 7.5\%, the precision and recall rates increased by 7.74\% and 6.94\%, respectively, and the overall accuracy was as high as 99.29\%. This result shows that the interaction between SU-FDE, SA, and DU has no conflict and significantly improves the model's overall performance in remote sensing image change detection.

\subsubsection{Effect of diffusion step size}

\begin{table}[!t]
\caption{The choice of diffusion step \( t \) changes the performance of CD on the CDD dataset}
\label{tab:CDD dataset about step t}
\begin{center}
\resizebox{\columnwidth}{!}{
\begin{tabular}{ccccc}
\hline
step \( t \) & F1-score & Precision & Recall & OA  \\ \hline
500    & 83.77 &  84.82  &  83.94  & 99.06     \\
750    & 85.11 &  85.31  &  85.10  & 99.10     \\
1000   & 88.47 &  89.13  &  87.35  & 99.29      \\\hline
\end{tabular}
}
\end{center}
\end{table}

We conducted ablation experiments to understand the impact of the time step \( t \) in the denoising diffusion model on the feature extraction capability for Change Detection (CD) datasets. During the training phase of the model, we considered multiple step settings to determine the appropriate time step length, including values of \( t \) set to 500, 750, and 1000. These experiments aimed to observe how different step length settings affect the model's performance on the CDD dataset. Table Six provides a detailed breakdown of the model's F1 scores, precision, recall rates, and overall accuracy at varying step lengths. Upon examination of the results, it was noted that the model achieved high values in all the aforementioned metrics at \( t = 1000 \). This observation in our model, which tracks CD over time, allows for a more detailed capture of changes in data details. The choice of step size t is crucial to the model's ability to process RS image data. The model can perform CD tasks more effectively under the setting of \( t = 1000 \).

\section{Conclusion}\label{sec:con}
In this study, we proposed the SMDNet model by introducing a Siamese network to propose a combination of SU-FDE and a denoising diffusion implicit model. We successfully applied it to the change detection task of high-resolution remote sensing images. Experimental results show that SMDNet's change detection performance on LEVIR-CD, DSIFN-CD, and CDD data sets has been significantly improved compared with existing technologies. Use the denoising capabilities of the diffusion model and the advantages of capturing data distribution. Add SU-FDE further to enhance the model's capture of edge details, thereby achieving more accurate change detection in complex scenes.
Despite the good performance of this model, there are still limitations.

1. In RS, due to the large amount of data and many resolutions from low to high, if you want to obtain more information, this will increase the time and cost of training, and it is difficult to implement in practice. Therefore, how to construct lightweight models to reduce training costs is an important direction.

2. We will continue to pay attention to the impact of the attention mechanism on the model. Explore an attention mechanism more suitable for RS images and conduct further research.

3. Considering the rapid advancements in RS hardware, a substantial volume of unlabeled data remains underutilized. Manually labeling data is time-consuming and often requires specific prior knowledge, yet it is still prone to inaccuracies such as under-labeling or mislabeling. Therefore, it is necessary to explore semi-supervised or self-supervised methods to leverage more data.

Future research will focus on further optimizing the model structure to reduce training costs, improve robustness under complex environmental conditions, and explore the potential of the SMD model in processing tasks in unlabeled remote sensing images.

\bibliographystyle{IEEEtran}
\bibliography{MyBiB}

\end{document}